\documentclass[conference]{IEEEtran}
\usepackage{times}

\usepackage{xcolor} 
\usepackage[numbers]{natbib}
\usepackage{multicol}
\usepackage[bookmarks=true]{hyperref}
\usepackage{amsmath}
\usepackage{booktabs}
\usepackage{tabularx}
\usepackage{ragged2e}
\usepackage{array}
\usepackage{graphicx}
\usepackage{algorithm}
\usepackage{algpseudocode}
\usepackage{tikz}
\newcolumntype{Y}{>{\RaggedRight\arraybackslash}X}
\newcolumntype{N}{>{\RaggedRight\arraybackslash}p{2.2cm}} 

\begin{document}

\title{A Modern System Recipe for Situated Embodied Human–Robot Conversation with Real-Time Multimodal LLMs and Tool-Calling}

\author{Author Names Omitted for Anonymous Review. Paper-ID 420}



%
\author{\authorblockN{Dong Won Lee\authorrefmark{2},
Sarah Gillet\authorrefmark{2},
Louis-Philippe Morency\authorrefmark{3}, 
Cynthia Breazeal\authorrefmark{2} and
Hae Won Park\authorrefmark{2}}
\authorblockA{\authorrefmark{2}Massachussetts Institute of Technology}
\authorblockA{\authorrefmark{3}Carnegie Mellon University}}


\maketitle

\begin{abstract} 
Situated embodied conversation requires robots to interleave real-time dialogue with active perception: deciding \emph{what to look at, when to look, and what to say} under tight latency constraints. We present a simple, minimal system recipe that pairs a real-time multimodal language model with a small set of tool interfaces for attention and active perception. We study six home-style scenarios that require frequent attention shifts and increasing perceptual scope. Across four system variants, we evaluate turn-level tool-decision correctness against human annotations and collect subjective ratings of interaction quality. Results indicate that real-time multimodal large language models and tool use for active perception is a promising direction for practical situated embodied conversation.

\end{abstract}


\IEEEpeerreviewmaketitle

\section{Introduction}
\label{sec:intro}
Enabling robots to \emph{converse naturally while jointly attending to a shared physical world} with users remains a central goal of human–robot interaction. Consider a simple exchange at home: a user asks, ``Where should I put this lamp?'', then the robot gazes at the lamp, inspects it, and turns to the shelf on the other side of the wall and says, "How about next to that shelf far away from the window?" In this scenario, the robot must (i) manage interaction timing (turn-taking, interruptions) to have a smooth exchange with the user, (ii) ground language in the currently visible scene to make a suggestion or, and (iii) plan its own sensing actions, e.g., decide if it needs additional visual information, where to look, whether to search outside of its current field of view for the shelf, and when it should look back at the speaker. In situated embodied conversation, the robot must tightly couple dialogue and perception  to allow for a smooth exchange with no perceivable delays and an awareness of the surroundings important for grounding and maintaining shared attention.

Various research communities have worked on realizing different elements that allow for \textbf{situated embodied conversation}. While a rich body of prior work has explored fluent dialogue and interaction timing in spoken HRI (e.g., turn-taking and dialogue management) \cite{skantze2021turntaking,marge2022spoken,reimann2024dmhri,gervits2020turnentry}, shared attention and joint attention \cite{scassellati1996sharedattention,breazeal1999attention,kontogiorgos2017multimodal,huang2010jointattention,pereira2019responsive,pereira2020contexts}, and active perception, grounding for situated language understanding \cite{lemaignan2012grounding,ognibene2022activevision,robinson2023roboticvision,cohen2024languagegrounding}, there is still no widely adopted systems abstraction for tightly interleaving real-time conversation with self-directed sensing actions across tasks, environments, and robot platforms. 

The emergence of real-time multimodal language models (Openai Realtime \cite{openai_realtime_api_reference}, Gemini Live \cite{google_gemini_live_get_started}) fundamentally changes this design space. Unlike prior dialogue systems, these models can generate language while continuously consuming audio and visual input. However, this capability alone is insufficient for situated embodied conversation. In physical interaction, a robot must also actively ground language in the environment and manage shared attention through self-directed sensing actions, i.e., deciding what to look at, when to look, and how to coordinate perception with ongoing dialogue. Therefore, “plugging a real-time model into a robot” is not enough -- without mechanisms for active grounding and attention control, it leaves a critical gap between fluent speech and situated understanding. A promising direction forward is tool use (a.k.a. function calling). Recent real-time LLMs provide this core functionality to emit structured function calls with typed arguments that a robot system can execute synchronously while the conversation continues.

This paper presents a minimal but expressive system recipe for situated embodied conversation, identifying the necessary and sufficient coordination mechanisms between real-time language generation and active perception. Our system is organized around two components: (1) a real-time multimodal LM that serves as a dialogue manager, (2) a set of tools provided to the LM that allows for directing the robots attention, i.e., gaze, and actively requesting new visual information. In sum, the realtime LM decides \emph{what to say, what to look at, and when} by interleaving conversation generation with structured actions such as \texttt{look\_at\_person}, \texttt{look\_at\_object}, \texttt{look\_around}, \texttt{look\_for}, and \texttt{use\_vision}.

We study this recipe across six home-style scenarios that require shifting attention between the interlocutor and task-relevant objects with increasing demands on perceptual scope: posture coaching, whiteboard tutoring, lamp placement, plant diagnosis, outfit checking, and finding misplaced items. We further evaluate four system variants via controlled ablations and backend comparisons, combining turn-level human annotations of ``what the robot should do next'' with subjective measures of fluency and social presence. This evaluation isolates which ingredients most strongly affect perceived situatedness and where current real-time LMs still struggle.

In summary, this paper makes the following contributions:
\begin{itemize}
    \item A system architecture pairing a real-time multimodal LM with tool-mediated attention and active perception for situated conversation.
    \item A set of tool interface for embodied conversation, including on-demand vision queries, person/object fixation, exploratory scene sweeps, and query-driven view retrieval, plus a lightweight viewpoint ``memory map'' for off-camera grounding.
    \item An evaluation protocol and scenario suite for situated embodied conversation, combining objective turn-level tool-call correctness with human questionnaires under controlled ablations and realtime model backends.
\end{itemize}

\begin{figure*}[t]
\centering
\includegraphics[width=1\linewidth]{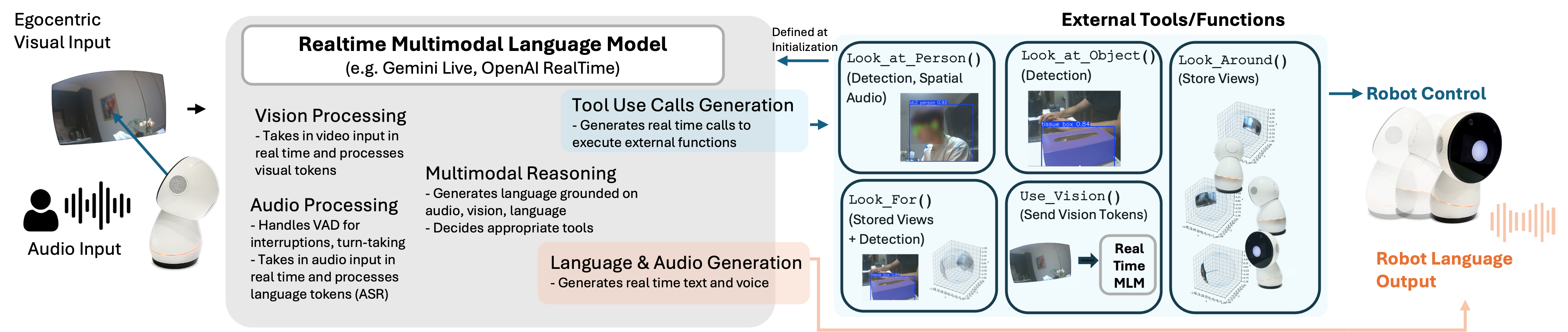}
\vspace{-0.4cm}
\caption{Overview of our real-time situated conversation system. Streaming egocentric vision and audio are processed by a real-time multimodal LM, which (i) generates spoken dialogue and (ii) issues low-latency function calls to external tools for attention and active perception (e.g., \texttt{Look\_at\_Person, Look\_at\_Object, Look\_Around, Look\_For, Use\_Vision}). Tool outputs update the shared perceptual context and drive robot gaze.}
\vspace{-0.3cm}
\label{fig:main}
\end{figure*} 
\section{Related Work}
\label{sec:related}

\subsection{Situated Human-Robot Conversations}
We use \emph{situated conversation} to mean dialogue whose meaning and interaction flow are conditioned on a \emph{shared, perceivable environment} (objects, people, visual understanding, and ongoing activities). In robotics and HRI, situated dialogue systems explicitly connect incremental language understanding to perceptual context, maintaining and revising candidate interpretations as the robot observes the world and as the world changes \cite{kruijff2007incremental,kruijff2006reference,zender2009refexp, lemaignan2012grounding}, where architectures bidirectionally couple situation awareness and dialogue processing \cite{kruijff2007incremental,kruijff_lison_book}. Because embodied sensing is viewpoint-limited, agents must actively select \emph{where to look} to acquire task-relevant evidence \cite{bajcsy2018revisiting}. In situated conversation, attention is additionally constrained by interactional demands (turn-taking, interruptions, addressee selection) and social norms (e.g., gaze signaling), making attention both an information-gathering policy and a communicative action \cite{mutlu2009footing,admoni2017gaze}. Real-time multimodal LMs have the potential to address this gap by treating situated conversation as an \emph{online} process: the agent can track streaming audio/video, react to interruptions, and revise referents as the world changes. However, closing this loop  \emph{requires active perception} the robot must move its sensors to acquire visual information  and to demonstrate attention.

\subsection{Real-Time Multimodal Large Language Models (LLMs)}

A recent and rapidly emerging direction is \emph{real-time} multimodal LLM systems that support low-latency, bidirectional interaction over continuous streams of speech and (increasingly) live visual context. In contrast to turn-based multimodal models that consume static images or pre-recorded clips, real-time systems such as production-grade interfaces such as OpenAI’s Realtime API \cite{openai_realtime_api_reference} and Google’s Gemini Live API \cite{google_gemini_live_get_started} are designed around streaming I/O, incremental state updates, and tight latency constraints, which expose persistent session abstractions that accept ongoing audio (and optionally image/video) inputs and generate interleaved audio/text outputs suitable for natural conversation.

\subsection{LLM Tool-Use and Function Calling}

A complementary line of work equips LLMs with \emph{tool-use}: the ability to invoke external functions (e.g., web search, databases, robot controllers, code execution) and incorporate the returned results into subsequent generation. Conceptually, tool-use frames the LLM as a high-level policy over an \emph{action space} of tools, where each action is a structured API call with arguments \cite{yao2022react}, recent progress has moved from \emph{ad hoc} text-form action formatting to \emph{constrained, typed} interfaces that improve reliability and integration into real systems, models have been trained to decide \emph{whether} to call tools and \emph{how} to format calls, including self-supervised tool-use training (Toolformer) \cite{schick2023toolformer} and API-call specialization with retrieval over tool documentation (Gorilla) \cite{patil2024gorilla}. Industrial APIs have largely standardized tool-use via JSON-schema-like function signatures and structured outputs, enabling developers to expose application actions as callable tools with argument validation (e.g., OpenAI function/ Gemini tool calling) \cite{openai_realtime_api_reference, google_gemini_live_get_started}. Overall, tool-use has matured into a practical mechanism for grounding LLM behavior in external state and actions, and it is now a core ingredient in embodied systems (as in our robot attention and perception tools).

\subsection{LLM integration for Human Robot Embodied Conversation} 

A growing set of HRI systems integrate LLMs as the dialogue core for embodied social robots, but the integration varies by application context and system depth. In \emph{open-domain and mixed-initiative social interaction} settings, LLMs are typically embedded as the central dialogue manager driving multi-party \cite{addlesee2024multi} or front-desk style interaction \cite{cherakara2023furchat} and, in some cases, longitudinal conversation over repeated encounters \cite{mauliana2025exploring,pinto2025designing}. In \emph{socially assistive robotics} for clinical \cite{blavette2025integrating} and elder-care contexts \cite{lima2025promoting}, LLMs are commonly integrated into existing SAR dialogue pipelines to improve conversational coverage and responsiveness. Complementing these systems, several works explore LLM-driven conversation on widely deployed commercial platforms such as Pepper/NAO in public-facing \cite{herath2025first} or educational scenarios \cite{smit2024enhancing}, often focusing on interaction quality and first-impression measures under constrained embodiment and sensing . Despite this progress, explicitly end-to-end streaming integrations with modern realtime multimodal APIs remain comparatively rare in the academic HRI literature. Moreover, the complementary line of work that uses pretrained LLMs for \emph{function calling}, i.e., invoking a discrete library of robot skills and perception tools, has largely been developed for task-centric manipulation/navigation \cite{ahn2022can,huang2023instruct2act,huang2022inner,wang2023voyager, wang2024lami}, rather than as part of an always-on conversational loop. We bridge these threads by coupling realtime multimodal inference with a tool interface for \emph{active perception} and \emph{attention control}, enabling the agent to interleave dialogue with actions to gather evidence and externalize attention during ongoing interaction.

\section{Methods}
\label{sec:method}

We implement a system for \emph{situated conversation}, where language is complemented with the robot’s perception of a shared environment embedded in real-time interaction flow (interruptions, turn-taking, and gaze shifts). The system is organized around two main components: (1) a real-time multimodal language model (LM) that handles streaming audio/video and conversational mechanics, (2) function calling for tool-mediated robot actions (primarily attentional control) including a lightweight ``memory map'' that stores surrounding views and queries them with a vision-language model (VLM) by using a geometric binding layer that maps video frames into the robot’s field-of-view (FOV) using rigid transformations. This design establishes a ``dialogue manager + tool use'' recipe in which a base multimodal LM is paired with perception and attention control.

\subsection{\textit{Real-time multimodal LM for streaming dialogue and visual grounding}}
\textit{Streaming inputs:} The robot provides continuous microphone audio and an egocentric camera stream. The real-time multimodal LM (e.g., OpenAI Realtime / Gemini Live) ingests these streams and maintains an online conversational state. \textit{Conversational mechanics (VAD and turn-taking):} We rely on the realtime LLM runtime to handle core dialogue mechanics, voice activity detection (VAD), turn-taking, and streaming ASR, so the system can support natural interruptions and quick turn exchanges. In practice, VAD boundaries define ``user turns'' that trigger downstream reasoning and tool decisions or interrupt ongoing robot actions. \textit{Multimodal grounding:} The LM can be periodically provided the most recent camera frame (or short frame bursts) during the conversation, enabling it to answer visual questions, and decide when additional visual evidence is required.

\subsection{\textbf{Function calling for robot attention control}}
Real-time multimodal LLM APIs encourage \emph{tool-augmented} interaction loops, i.e., sessions can incorporate structured events (e.g., function calling / external actions), allowing the LLM to interleave perception queries and attention control outputs with dialogue, providing a practical substrate for embodied or interactive agents that must coordinate speech, perception, and action in real time. 
We delegate tool selection to the pretrained LM. Given the live conversational context and the current perceptual state, the LM decides what action to take, and when, i.e., in which conversational turn, to do so (e.g., verify a reference by looking, command the robot system to search for an object  out of view, or to maintain face attention during turn-taking). This offloads a traditionally hand-engineered decision layer, often implemented as explicit rules or a learned policy, to the LM’s broad priors about social interactions.

\textit{Tool interface.} We expose robot capabilities and perception queries as  functions that the LM can call during dialogue. These functions form the bridge between language and embodied action. To enable situated embodied conversations, our primary tool family targets the ``where to look'' and ``when to look'' subproblems of interaction via a small set of attentional and perception primitives. In total, we implement five basic functions: \texttt{look\_at\_me}, \texttt{look\_at\_object}, \texttt{look\_around}, \texttt{look\_for}, and \texttt{use\_vision}. The first four functions support continuous attention control (tracking the interlocutor or a referenced object), exploratory sensing (scanning the scene), and targeted search (locating a specified person/object). Finally, \texttt{use\_vision} provides an explicit, on-demand perception query: rather than continuously streaming video to the LM, which can be costly as image tokens accumulate \ref{tab:cost_latency_and_pricing}, the system can request visual information to be sent to the realtime multimodal language model only when it is necessary, enabling more efficient and deliberate visual reasoning.

\begin{algorithm}[t]
{\small
\caption{\texttt{use\_vision}: send view to realtime LM}
\label{alg:send_latest_frame}
\begin{algorithmic}[1]
\Function{use\_vision}{$ws,\; q$}
  \Comment{$ws$: realtime websocket; $q$: text query/prompt}
  \State $I \gets$ latest camera frame
  \State send message to realtime LM containing $I,\; q$ 
\EndFunction
\end{algorithmic}
}
\end{algorithm}

\begin{algorithm}[t]
{\small
\caption{\texttt{look\_at\_person}: follow the person }
\label{alg:look_at_me}
\begin{algorithmic}[1]
\Function{look\_at\_person}{enabled}
  \While{enabled}
    \State $I \gets$ latest camera frame
    \State $(T, \mathrm{fov}) \gets$ latest robot pose / field-of-view
    \State $r \gets$ person tracker on $I$
    \State $A \gets$ latest spatial audio location
    
    \If{$r$ has keypoint}
      \State $\Delta(\mathrm{yaw},\mathrm{pitch}) \gets$ offset from image center
      \State $(x,y,z) \gets$ 3D gaze target from $(T,\mathrm{fov},\Delta)$
      \State publish gaze target $(x,y,z)$
    \Else
      \State $(x,y,z) \gets$ spatial audio estimate from A
      \State publish gaze target $(x,y,z)$
    \EndIf
  \EndWhile
\EndFunction
\end{algorithmic}
}
\end{algorithm}

\texttt{look\_at\_person} is a closed-loop attentional primitive that continuously steers the robot's gaze toward the speaker while it is enabled. On each iteration, it (i) reads the latest egocentric camera frame $I$ and the robot's current pose and camera field-of-view $(T,\mathrm{fov})$, then (ii) runs a person tracker to find interlocutors in the current camera image and use the extracted keypoints to determine the rotation $\Delta(\mathrm{yaw},\mathrm{pitch})$ needed to center the robot's attention on them. We complement visual tracking  with coarse spatial-audio-based speech localization for cases of tracking or detection failures, e.g., when the person is occluded or out of camera view. 
\emph{We utilize \texttt{look\_at\_person} as a default behavior when no other function calls are triggered by the model}. \texttt{look\_at\_object} mirrors \texttt{look\_at\_me}, replacing the person tracker with an object-specific perception module (object detection + temporal tracking). \footnote{For \texttt{look\_at\_person}, the exact target coordinate for the robot's attention is derived from a specific keypoint, e.g., the nose keypoint. For \texttt{look\_at\_object}, the target attention coordinate is calculated using the mask centroid or bounding-box center.} A key design goal is to keep the attentional control loop \emph{fast, local, and continuously reactive}: the robot must update gaze at frame-rate timescales. Therefore, we implement \texttt{look\_at\_person} and \texttt{look\_at\_object} as lightweight attentional primitives that run on-device and expose only compact targets (e.g., keypoints or centroids) to the controller; In our work, we locally run \texttt{YOLO11-pose}~\cite{khanam2024yolov11} for person tracking, with mean inference time $49.3 \pm 3.50$\,ms (20.37\,Hz), and \texttt{SAM3}~\cite{carion2025sam} for object mask estimation/tracking, with mean inference time $161.7 \pm 4.31$\,ms (6.18\,Hz), enabling real-time use.



\begin{algorithm}[t]
{\small
\caption{\texttt{look\_around}: sweep \& save views}
\label{alg:look_around}
\begin{algorithmic}[1]
\Function{look\_around}{$C,\; D$}
  \Statex \textit{$C$: list of look targets; $D$: output directory}
  \For{\textbf{each} target $\mathbf{p}_i=(x_i,y_i,z_i)\in C$}
    \State publish look-at $\mathbf{p}_i$
    \State wait until robot finishes motion
    \State $I_i \gets$ latest camera frame
    \State $(T_i,\mathrm{fov}_i) \gets$ latest pose / field-of-view
    \State save tuple $(I_i, T_i, \mathrm{fov}_i, \mathbf{p}_i)$ to $D$
  \EndFor
\EndFunction
\end{algorithmic}
}
\end{algorithm}

\begin{algorithm}[t]
{\small
\caption{\texttt{look\_for}: select most appropriate view from saved viewpoints for a text query then feed results to the realtime API}
\label{alg:find_in_world}
\begin{algorithmic}[1]
\Function{look\_for}{$q,\; D$} 
  \Statex \textit{$q$: text query given from LLM function call; $D$: directory where images from look\_around() are located}
  \State $\mathcal{V} \gets \{(I_i,\mathbf{p}_i)\}$ loaded from saved tuples in $D$
  \State \textbf{in $|V|$ concurrent threads}: 
  \State \hspace{\algorithmicindent} score each $(I_i,\mathbf{p}_i)\in\mathcal{V}$ with a VLM($I_i, q$) to  \State \hspace{\algorithmicindent} obtain relevance score $s_i\in[0,1]$
  \State $i^* \gets \arg\max_i s_i$
  \State publish look-at $\mathbf{p}_{i^*}$
  \State send message to realtime LM containing $I_{i^*}$ and $q$ 
\EndFunction
\end{algorithmic}
}

\end{algorithm}


\begin{figure}[t]
\centering
\includegraphics[width=0.8\linewidth]{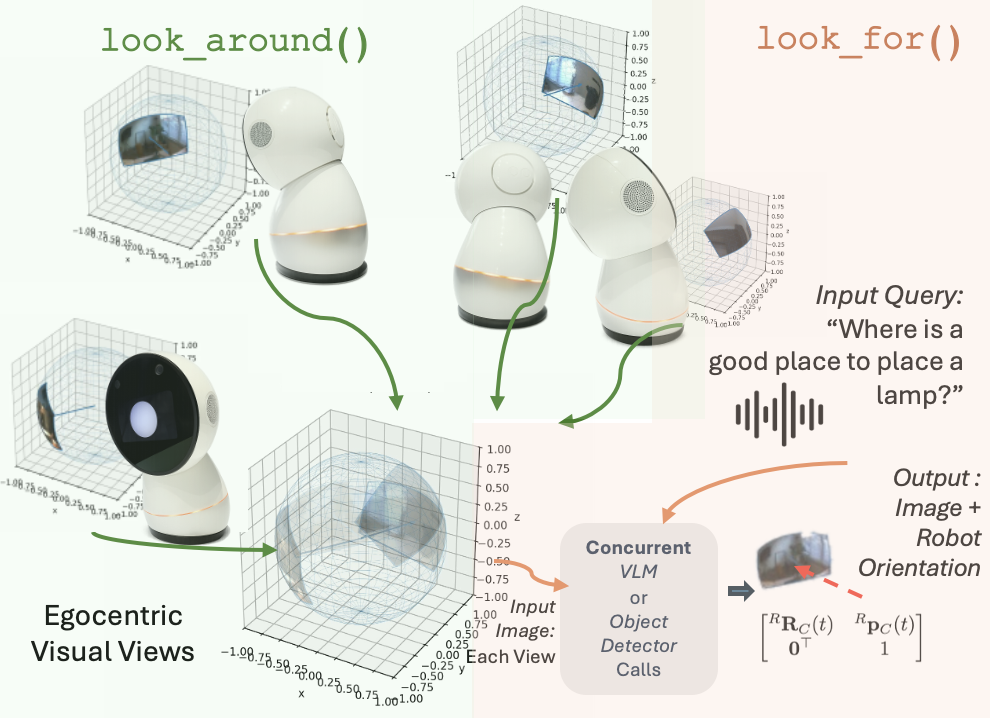}
\vspace{-0.4cm}
\caption{\texttt{look\_around} performs a sweep to acquire and store egocentric views with associated robot poses, forming a lightweight view-memory. Given a language query (e.g., “Where is a good place to place a lamp?”), \texttt{look\_for} searches over stored views using concurrent VLM/object-detector calls and returns the best-matching image evidence along with the corresponding robot orientation/pose for action.}
\vspace{-0.3cm}
\label{fig:viz_map}
\end{figure} 
\begin{figure*}[t]
\centering
\includegraphics[width=1\linewidth]{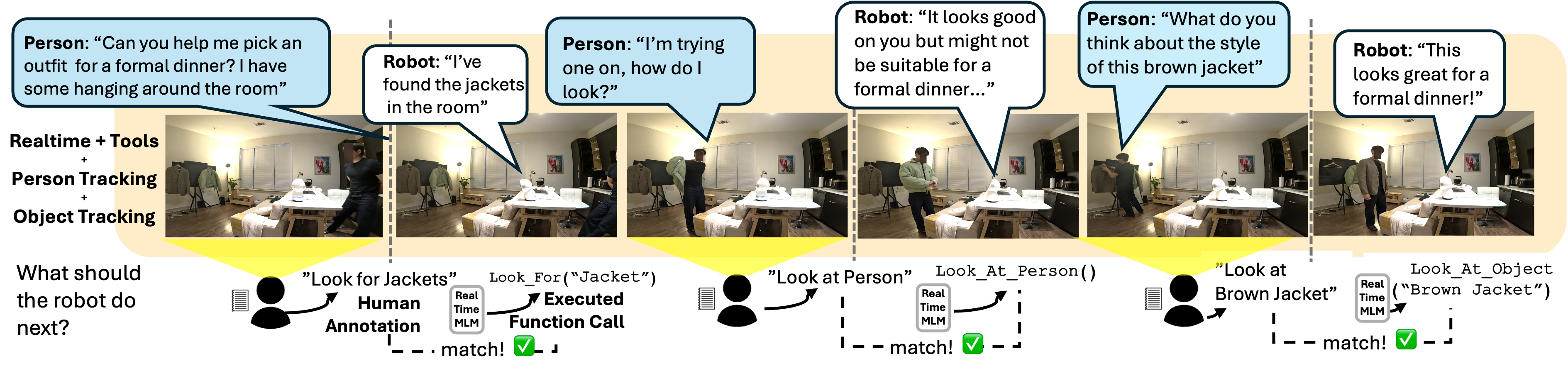}
\vspace{-0.4cm}
\caption{Example interaction illustrating turn-level “what should the robot do next?” tool-decision evaluation in an outfit-selection scenario. As the user’s dialogue evolves (finding jackets → trying one on → asking about a specific brown jacket), the real-time system interleaves spoken responses with  function calls. Human annotations specify the expected next perception action at each turn, enabling direct comparison to the executed tool calls.}
\vspace{-0.3cm}
\label{fig:anot}
\end{figure*} 
\texttt{look\_around} is an exploratory sensing primitive that performs a short, structured sweep of the environment and records a set of anchored viewpoints, i.e., images plus their position of capturing, for later retrieval. On the other hand, \texttt{look\_for} is a targeted retrieval-and-attend primitive outlined in Algorithm \ref{alg:find_in_world} that selects the most relevant previously observed viewpoint for a natural-language query. Given a query string $q$ (typically produced by an LM tool call) and a image directory $D$ populated by \texttt{look\_around}, the method 
evaluates each candidate viewpoint using a vision-language model, computing a relevance score $s_i=\mathrm{VLM}(I_i,q)\in[0,1]$ that measures how well the image supports the query. We parallelize VLM scoring to improve throughput and reduce end-to-end query time. The highest-scoring viewpoint is selected as the best evidence for $q$. The robot then directs its attention to replicate the position in which the viewpoint was captured. 
Finally, the selected frame is sent back to the real-time LM alongside $q$, allowing the dialogue manager to ground its response in the most informative stored observation. 

\noindent\textbf{Geometric Binding Layer}
To make egocentric video observations actionable for gaze control, we bind each camera frame to the robot’s camera pose and field-of-view (FOV) using a rigid transform ${}^{R}\mathbf{T}_{C}(t)\in SE(3)$ obtained from kinematics/IMU. This establishes a simple mapping from image-space targets (e.g., a pose keypoint, box center, or mask centroid) to robot-frame look-at commands: we back-project the target pixel to a camera ray using the camera intrinsics (or FOV), rotate that ray into the robot frame, and parameterize a 3D look-at point. The resulting $(x,y,z)$ goal can be published directly to the robot’s gaze controller, enabling closed-loop recentering for both people (\texttt{look\_at\_person}, using a face/nose keypoint) and objects (\texttt{look\_at\_object}, using a centroid/box center) with a shared geometric interface. Full derivations (pixel-to-ray, $\Delta(\mathrm{yaw},\mathrm{pitch})$ error, and look-at point construction) are provided in Appendix~\ref{app:geometric_binding}.

\section{Experiments}
\label{sec:exp}
In order to investigate the effectiveness of the proposed system, we designed six interaction scenarios that progressively increase the demands on visual grounding and perceptual scope. Furthermore, we tested four system variants to isolate the contribution of key system elements. For controlled comparison, we captured videos of a person interacting with the system for every variant, scenario combination, yielding 24 videos ($6\times4$), which are provided in the appendix. For each video, we collected expert human annotations of (i) \textit{system correctness}, 
and (ii) third-person observer judgments of the interaction quality. 


\begin{table}[t]
\centering
\footnotesize
\setlength{\tabcolsep}{2.0pt}
\renewcommand{\arraystretch}{1.10}
\resizebox{\columnwidth}{!}{%
\begin{tabular}{@{}p{2.15cm} p{5.35cm} p{1.25cm} c c c c@{}}
\toprule
\textbf{Task} & \textbf{Goal} & \textbf{Duration} & \textbf{Viz.} & \textbf{Person} & \textbf{Object(s)} & \textbf{Map} \\
\midrule
Posture coach &
Identify posture issue; coach and verify improvement. &
01:39 &
Y & Y (1) & N & N \\
Whiteboard &
Solve a trig problem using whiteboard diagram &
02:12 &
Y & Y (1) & Y (1) & N \\
Lamp placement &
Choose a lamp location using a room sweep. &
02:05 &
Y & Y (1) & Y (1) & Y (1) \\
Plant doctor &
Find plant and diagnose plant health. &
02:12 &
Y & Y (1) & Y (1) & Y (1) \\
Outfit check &
Find outfits in room, compare options on-body. &
02:21 &
Y & Y (1) & Y (2) & Y (1) \\
Pack / find objects &
Locate items via a scan; re-scan after changes. &
02:03 &
Y & Y (1) & Y (3) & Y (2) \\
\bottomrule
\end{tabular}%
}
\caption{Task descriptions and visual requirements: \textbf{Viz.} indicates whether the task requires visual input.
\textbf{Person}, \textbf{Object(s)}, and \textbf{Map} specify which visual entities are required. Counts in parentheses give the number of distinct targets or map updates/sweeps.}
\label{tab:task_taxonomy}
\end{table}

\subsection{Interaction Scenarios}
Table \ref{tab:task_taxonomy} outlines the tasks and how the different tasks require different parts of the abilities needed for situated embodied conversation. The simpler tasks require local, single-target understanding (e.g assessing a person's posture or tutoring instructions with a single diagram), while more complex tasks require room-scale visual context, multi-object reference and scene changes that force the agent to actively gather evidence (e.g finding lamp placement location, choosing outfits distributed in the scene, or locating multiple objects via repeated scans after the user modifies the environment). This suite of 6 tasks form a mini-testbed to probe our method's robustness to varying levels of complexity, from person-only, to person + single object, to person + single object + surrounding understang, to person + multiple objects + multiple changes in the surrounding. These six tasks are a representative, author-selected subset; expanding to a wider set of scenarios is left to future work.

\subsection{System Ablations}

We designed four different system variations to understand the performance and contribution of the different system elements, namely:

\begin{itemize}
\item \textbf{OpenAI Realtime \cite{openai_realtime_api_reference}:} our proposed system with person, object, and environment sweep tools enabled.
\item \textbf{OpenAI Realtime w/o Object:} disables object tracking and search, specifically  \texttt{look\_for} and \texttt{look\_at\_object}
\item \textbf{OpenAI Realtime w/o Person:} disables person tracking: \texttt{look\_at\_person} 
\item \textbf{Gemini Live \cite{google_gemini_live_get_started}:} an alternative real-time multimodal backend with all tools enabled. This system variant tests backend \emph{swapability}: by holding the embodiment/tool layer fixed and changing only the streaming multimodal model.
\end{itemize}

We evaluated these four different variations with the same scenarios to explore generalizability and limitations of the different variants and our proposed system with captured videos of a person interacting with the system in every variant and scenario combination.

\subsection{Human Annotations: System Correctness Measures}
As the realtime LLMs operate on conversational turns, we asked annotators to provide annotations for what the robot should do during the next turn exchange. For each turn exchange, annotators labeled whether the robot should (1) look for an object (analogous to \texttt{look\_for}), (2) should fix gaze on an object (\texttt{look\_at\_person}) or (3) on the person (\texttt{look\_at\_object}), as shown in Fig. \ref{fig:anot}. In addition, annotators indicated whether the robot needed to use visual understanding to respond correctly at that moment, that is, whether the next response depends on the current visual scene rather than language alone (analogous to \texttt{use\_vision}). For example, when the person asks ``Does this plant look healthy?'' the robot needs to consider the current video feed to answer this question and continue the conversation. Using these labels, we compare the model's decision against the human reference to compute accuracy, precision and recall for each system variant and overall macro-average. For annotator consistency, two annotators first coded two videos and after discussing alignment coded the remaining videos. An interrater reliability measured from four videos on the ``should-do'' behavior revealed a Cohen's kappa score of 0.41, which indicates fair to moderate agreement. Closer inspection showed that annotators disagreed when \texttt{look\_for} vs. `\texttt{look\_at\_object} should be used.

\begin{table*}[t]
\centering
\small
\setlength{\tabcolsep}{4pt}
\resizebox{\textwidth}{!}{%
\begin{tabular}{lccc ccc ccc ccc ccc}
\toprule
& \multicolumn{3}{c}{Overall} 
& \multicolumn{3}{c}{\texttt{look\_at\_person()}}
& \multicolumn{3}{c}{\texttt{look\_at\_object()}}
& \multicolumn{3}{c}{\texttt{look\_for()}}
& \multicolumn{3}{c}{\texttt{use\_vision()}} \\
\cmidrule(lr){2-4}\cmidrule(lr){5-7}\cmidrule(lr){8-10}\cmidrule(lr){11-13}\cmidrule(lr){14-16}
Condition & Acc. & P & R & Acc. & P & R & Acc. & P & R & Acc. & P & R & Acc. & P & R \\
\midrule
OpenAI Realtime \cite{openai_realtime_api_reference}         & $0.72 \pm 0.10$ & $0.88 \pm 0.18$ & $0.62 \pm 0.23$ & 0.63 & 0.52 & 1.00 & 0.70 & 1.00 & 0.19 & 0.93 & 1.00 & 0.70 & 0.64 & 1.00 & 0.60 \\
OpenAI Realtime \cite{openai_realtime_api_reference} \textit{w/o Object}  & $0.28 \pm 0.28$ & $0.50 \pm 0.50$ & $0.26 \pm 0.26$ & 0.42 & 1.00 & 0.42 &      &      &      &      &      &      & 0.70 & 1.00 & 0.64 \\
OpenAI Realtime \cite{openai_realtime_api_reference} \textit{w/o Person} & $0.48 \pm 0.24$ & $0.54 \pm 0.27$ & $0.35 \pm 0.18$ &      &      &      & 0.69 & 0.62 & 0.40 & 0.71 & 0.53 & 0.53 & 0.51 & 1.00 & 0.48 \\
Gemini Live \cite{google_gemini_live_get_started}        & $0.77 \pm 0.07$ & $0.84 \pm 0.16$ & $0.60 \pm 0.22$ & 0.76 & 0.68 & 1.00 & 0.76 & 0.67 & 0.17 & 0.91 & 1.00 & 0.60 & 0.65 & 1.00 & 0.62 \\
\bottomrule
\end{tabular}%
}
\caption{System Correctness Measures: LM tool-call predictions compared to human “should-do-next” annotations. We report macro-averaged accuracy (Acc.), precision (P), and recall (R) for each system condition, both overall and broken down by tool category \texttt{(look\_at\_person, look\_at\_object, look\_for, use\_vision)}}
\label{tab:metrics_by_condition}
\end{table*}

\subsection{Overall Interaction Quality Measures}
We used subjective measures in form of questionnaires to evaluate how the different variations of the system impacted user perception. We utilized the ``Attention allocation'' subscale of the Networked Minds Questionnaire~\cite{harms2004internal} and further evaluated the user's perception of the robot's conversational abilities through questions targeting conversational continuity, temporal fluency and turn-taking as well as understanding of referential wording. 

\noindent \textbf{Fluent and Smooth} These items capture the overall interactional flow, serving as a coarse, holistic check on perceived fluency. 

\begin{itemize}
    \item (1) The conversation felt smooth rather than effortful. 
    \item (2) The interaction felt fluent overall.
\end{itemize}

\noindent \textbf{Convo. Quality} This category evaluates the perceived competence of the underlying LLM that drives the real-time system—i.e., whether the robot’s utterances are coherent, contextually appropriate, and linguistically well-formed. It serves as a backend-quality control measure, separating “core language ability” from effects introduced by active perception.

\begin{itemize}
    \item (1) The robot’s responses related clearly to what the person just said.
    \item (2) The robot stayed on topic throughout the conversation.
\end{itemize}

\noindent \textbf{Attention to Person} Selected from the Networked Minds Questionnaire~\cite{harms2004internal}, these items probe whether the robot is perceived as socially attending to the interlocutor—detecting, acknowledging, and orienting toward the person in a way that signals awareness. 

\begin{itemize}
    \item (1) The person’s presence was obvious to Jibo.
    \item (2) The person caught Jibo’s attention.
\end{itemize}

\noindent \textbf{Reference to Object} These questions evaluate situated referential grounding: whether the robot correctly resolves references to objects/places in the shared environment and aligns its attention accordingly.  

\begin{itemize}
    \item (1) The robot understood what entity, object or space in the surroundings the person was referring to.
    \item (2) The robot understood what object the person was holding and referring to.
    \item (3) The robot directed its attention to the entity, object or space the person was referring to.
\end{itemize}

\noindent \textbf{Response Timing \& Turn Taking} These items capture perceived turn-transition naturalness, which is particularly relevant to end-of-speech detection and response onset latency in real-time systems. 

\begin{itemize}
    \item (1) The robot responded at appropriate times. 
    \item (2) The robot’s response timing felt natural.
\end{itemize}

\noindent \textbf{Interruption \& Pauses} These items capture coordination under overlap and silence, which is strongly influenced by start-of-speech detection and the system’s ability to yield the floor quickly. (1) The robot avoided interrupting the person. (2) The person could interrupt the robot when needed. (3) Pauses in the interaction felt appropriate.

\begin{itemize}
    \item (1) The robot avoided interrupting the person.
    \item (2) The person could interrupt the robot when needed.
    \item (3) Pauses in the interaction felt appropriate.
\end{itemize}

We worked with two human annotators (35F, 18F) to obtain annotations and subjective evaluations of the different system variations. Both annotators were knowledgeable about the general system functionality but blind to the system variations while annotating and filling out questionnaires. All items were rated on a 5-point Likert scale (1 = Strongly disagree, 5 = Strongly agree).  Annotators re-watched videos they had just coded for the system correctness annotations and also the remaining videos after annotations were done to fill the questionnaire for subjective measures.

\section{Results}
\label{sec:res}

\subsection{Evaluations on System Correctness Annotations}
We refer the reader to Table \ref{tab:metrics_by_condition}, where we display results for system correctness. Both full systems -- Gemini and OpenAI -- achieve similar turn-level tool-decision correctness (macro-accuracy), where “correct” means the model’s next-step perception tool call matches human “should-do-next” annotations for that turn: OpenAI Realtime attains $0.72\pm0.10$ macro-averaged accuracy across tool categories with $0.88\pm0.18$ precision and $0.62\pm0.23$ recall, while Gemini Live reaches $0.77\pm0.07$ accuracy with $0.84\pm0.16$ precision and $0.60\pm0.22$ recall. Here, precision measures agreement on tool invocations, when the model calls a tool, annotators also labeled that tool as needed for that turn, while recall measures coverage of annotated needs, when annotators labeled a tool as needed, the model actually calls it. In interaction terms, low precision corresponds to unnecessary attention shifts, whereas low recall corresponds to missed sensing. This suggests that \textbf{at this time}, swapping the real-time multimodal API preserves overall tool-decision quality, with modest differences in variance and precision/recall trade-offs.

We refer the readers to the right of Table~\ref{tab:metrics_by_condition}, where we find that \texttt{look\_for()} is the most reliable decision: both OpenAI and Gemini achieve near-perfect precision ($P=1.00$) with strong recall ($R=0.70$ and $0.60$, respectively) and high accuracy ($0.93$ and $0.91$). Second, \texttt{use\_vision()} also exhibits perfect precision ($P=1.00$) but only moderate recall ($R\approx0.60$), indicating the dialogue manager tends to request visual evidence conservatively, missing some turns where annotators judged that the next response truly depends on the current scene. For both OpenAI and Gemini, \texttt{look\_at\_person()} exhibits perfect recall ($R=1.00$), but lower precision ($P=0.52$ and $0.68$), consistent with a tendency to over-trigger person-attention behaviors (hard-coded safe social default, \ref{sec:method}) while rarely missing turns that require attending to the interlocutor.

\paragraph{Per-ablation Patterns}
The ablated system variants are expected to degrade in overall correctness because they remove the functions required to support the corresponding behaviors. OpenAI Realtime w/o Object removes the object tracking and search tools (\texttt{look\_for()}, \texttt{look\_at\_object()}). Within the remaining tool set, this system variant exhibits a sharper precision--recall trade-off for person attention: \texttt{look\_at\_person()} reaches perfect precision ($P=1.00$) but substantially lower recall ($R=0.42$), compared to the full system’s $P=0.52, R=1.00$. OpenAI Realtime w/o Person disables person tracking (\texttt{look\_at\_person()}). For the remaining object/search tools, we observe a relative shift toward object-directed behavior: \texttt{look\_at\_object()} recall increases (from $R=0.19$ to $R=0.40$) with reduced precision (from $P=1.00$ to $P=0.62$), while \texttt{look\_for()} degrades markedly (Acc.=0.71, $P=0.53$, $R=0.53$ vs.\ Acc.=0.93, $P=1.00$, $R=0.70$). The prompt-level descriptions of \texttt{look\_for()} and \texttt{look\_at\_object()} are linguistically similar, and both correspond to object-directed attention behaviors, therefore the model may fail to reliably discriminate between ``search'' versus ``fixate”. Consequently, this suggests that careful tool definition and prompt design  is important for robust tool selection.


\subsection{Evaluations on Overall Interaction Quality}
Tab.~\ref{tab:subjective_composites_meanstd} summarizes these category-level distributions across the six scenarios. For \emph{Convo. Quality}, we find that all variants perform consistently well with an average score of $4.65 \pm 0.15$ out of 5 across all system variants. In contrast, categories that directly reflect \emph{Attention to Person} and \emph{Object} seem to be sensitive to the ablations. Disabling person tracking perception (\textsc{OpenAI} w/o person) appears to reduce \emph{Attention to Person} and increase variance, consistent with a weaker sense that the robot notices and orients to the interlocutor. Disabling object tracking (\textsc{OpenAI} w/o objects) appears to produce the largest drop in \emph{Reference to Object}, indicating degraded ability to resolve and physically align attention to referents in the shared environment when object/search tools are unavailable. Importantly, \emph{Fluent and Smooth} are more holistic measures of interaction quality. We find that the full version of OpenAI and Gemini perform well, obtaining an average score of $4.62 \pm 0.43$ and $4.42 \pm 0.85$  respectively. Whereas, this score appears to drop for the ablated variants, calling for future work to investigate if attention behaviors are crucial for overall interaction quality. \emph{Interruption \& Pauses} and \emph{Response Timing \& Turn Taking} measure the underlying Voice Activity Detection proficiency of realtime systems. \emph{Interruption \& Pauses} is consistently high across all system variants (implying that the start of speech detection is generally adequate to handle basic overlap and silence management.) On the other hand, we find that \emph{Response Timing \& Turn Taking} shows larger spread for Gemini Live, which warrants further investigation into the end-of-speech detection behavior from Gemini Live's VAD. 

\begin{table}[t]
\centering
\label{tab:subjective_composites_meanstd}
\resizebox{\columnwidth}{!}{%
\begin{tabular}{lc|ccc}
\toprule
\textbf{Category} & \textbf{Gemini} & \textbf{OpenAI} & \textit{\textbf{w/o person}} & \textit{\textbf{w/o objects}} \\
\midrule
Fluent and Smooth              & $4.42 \pm 0.85$ & $4.62 \pm 0.43$ & $3.96 \pm 0.96$ & $3.79 \pm 0.92$ \\
Convo.\ Quality                & $4.79 \pm 0.33$ & $4.79 \pm 0.26$ & $4.50 \pm 0.60$ & $4.50 \pm 0.64$ \\
Attention to Person            & $4.29 \pm 1.18$ & $4.75 \pm 0.34$ & $4.12 \pm 1.03$ & $4.67 \pm 0.39$ \\
Reference to Object            & $4.61 \pm 0.55$ & $4.83 \pm 0.33$ & $4.08 \pm 1.09$ & $3.47 \pm 0.96$ \\
Response Timing \& Turn Taking & $3.89 \pm 1.35$ & $4.83 \pm 0.17$ & $4.92 \pm 0.15$ & $4.72 \pm 0.34$ \\
Interruption \& Pauses         & $4.94 \pm 0.19$ & $5.00 \pm 0.00$ & $4.97 \pm 0.10$ & $4.97 \pm 0.10$ \\
\bottomrule
\end{tabular}%
}
\caption{Overall Interaction Quality Results (mean $\pm$ std; $n = 12 \times m_c$ item ratings per condition, where $m_c$ is the number of survey questions in category $c$): We aggregate Likert responses into six  categories and report scores for each system condition}
\vspace{-0.3cm}
\end{table}

\begin{table}[t]
\centering
\small
\setlength{\tabcolsep}{4pt}
\resizebox{\columnwidth}{!}{%
\begin{tabular}{lcc}
\toprule
\multicolumn{3}{c}{\textbf{(A) Measured runtime cost and latency}} \\
\midrule
 & OpenAI Realtime \cite{openai_realtime_api_reference} & Gemini Live \cite{google_gemini_live_get_started} \\
\midrule
Cost (\$)    & $0.047 \pm 0.013$   & $0.018 \pm 0.002$ \\
Latency (ms) & $689.98 \pm 389.08$ & $739.10 \pm 593.45$ \\
\midrule
\multicolumn{3}{c}{\textbf{(B) API pricing by modality (USD / 1M Tokens)}} \\
\midrule
 & OpenAI Realtime \cite{openai_realtime_api_reference} & Gemini Live \cite{google_gemini_live_get_started} \\
\midrule
Inputs (Text/Audio/Image) & 4.00 / 32.00 / 5.00 & 0.50 / 3.00 / 3.00 \\
Outputs (Text/Audio)      & 16.00 / 64.00       & 2.00 / 12.00 \\
\bottomrule
\end{tabular}%
}
\caption{Runtime measurements (A) and API pricing (B). Pricing in USD per 1M tokens.}
\label{tab:cost_latency_and_pricing}
\end{table}

\subsection{Latency and Cost}
We compare the two real-time backends in terms of end-to-end interaction latency and per-interaction API cost under the same system stack and scenarios. In our runs, both backends exhibit similar latency on average (OpenAI Realtime: $690\pm389$ ms; Gemini Live: $739\pm593$ ms), with substantial variance that likely reflects natural heterogeneity in turn structure and streaming end-of-turn detection. In contrast, cost differs markedly: Gemini Live is substantially cheaper per interaction in our measurements ($\$0.018\pm0.002$) than OpenAI Realtime ($\$0.047\pm0.013$), reflecting different pricing of streaming modalities and output generation.

Overall, these results suggest that the choice of real-time multimodal backend is largely a \emph{deployment} knob: swapping OpenAI Realtime for Gemini Live preserves interaction timing at the scale of our measured end-to-end latency, but can change operating cost by a multiple, which becomes consequential for longer-running deployments or higher interaction volumes.

\section{Discussion}
\label{sec:discussion}


\subsection{Tool-mediated attention can be the main driver of situational awareness}
Firstly, the full systems (with person, object, and environment tools enabled) achieve similar turn-level tool-decision correctness across the two real-time backends. Future work should therefore investigate if the proposed embodiment/tool layer is truly  ``backend-agnostic'' beyond the two backends explored in this work. Second, subjective judgments remain near-ceiling for conversational coherence across system variants, while categories tied to embodiment---especially attention-to-person and reference-to-object---seem to drop under ablations that remove the corresponding tools. Taken together, these findings support the view that the practical benefits of real-time multimodal LMs for embodied interaction are mediated primarily through the agent's ability to \emph{act} (e.g., reorient attention, request visual evidence, or search the scene) at the right moments, rather than through improvements in language alone.

\subsection{Per-tool lessons: precision is easy; recall is the bottleneck}
Across backends, several tools are invoked correctly when invoked (high precision), yet are under-invoked (moderate/low recall). In particular, object fixation (\texttt{look\_at\_object}) shows very low recall despite high precision, while explicit vision requests (\texttt{use\_vision}) are conservative---often correct, but missing some turns where annotators judged visual evidence was needed. This is a common failure mode for tool-using dialogue managers: models learn a ``safe'' policy that avoids unnecessary calls, but in embodied interaction, \emph{not looking} is itself an error because it breaks common ground and forces users into repair behaviors. This suggests that future optimization should target \emph{calibration and coverage} (when to look) as much as correctness (how to look), potentially via costs that penalize missed perceptual actions, state features that expose perceptual uncertainty, or training data that includes repair-heavy episodes.

\subsection{Prompt and interface design matter more than it seems}
The \texttt{look\_for} and \texttt{look\_at\_object} tools are linguistically and functionally close: both are ``object-directed,'' but one implies retrieval from stored views while the other implies fixation in the live view. The observed confusion under the w/o-person ablation is consistent with a tool defintion that is underspecified or ambiguous from the model's perspective. A practical implication is that future systems should make tool boundaries mutually exclusive where possible, and return structured state.

\subsection{Applicable to other embodiments}
A key design goal of this work is that the overall recipe is \emph{robot-agnostic}. Porting the system to a new robot therefore primarily requires implementing a thin adapter that (a) streams microphone audio and an egocentric camera feed to the real-time backend, and (b) maps tool calls into the robot's available sensing and actuation (e.g., head yaw/pitch control and camera stream). Beyond Jibo (the robotic platform used in our study), we implemented the same recipe on a second platform (Reachy Mini) by reusing the identical real-time dialogue manager and tool schemas, and swapping only the embodiment adapter (camera/audio IO + gaze/attention control). We refer readers to the supplementary materials for cross-platform demonstration videos and qualitative comparisons.

\subsection{Limitations}
\paragraph{Turn-based system}
 A limitation of current realtime LLM systems is that operates at the turn level but embodied interaction is often driven by sub-turn phenomena: incremental speech, micro-adjustments of gaze, interruptions, and rapid perception updates. 
 Future work should incorporate more temporally granular actions and accompanying metrics (e.g., time-to-attend after a referential cue) that reflect the dynamic nature of embodied conversation.

\paragraph{Annotation subjectivity and measurement noise.}
Our evaluation relies on expert annotators and subjective measures of conversational quality and attentional appropriateness. Some judgments---especially about gaze---can be difficult from certain camera angles or when head pose cues are subtle, introducing noise and reducing apparent agreement. These ambiguities have affected our evaluations but occured in all system variants similarly. 

\paragraph{Scenario and user diversity}
We evaluate on six scenarios, which is limited from the standpoint of broad generalization, yet still meaningful given the combinatorial variability of situated interaction and the cost of producing high-quality annotated multimodal episodes. In practice, these scenarios offer a first glimpse into whether a single recipe transfers across qualitatively different home-style contexts, but they cannot establish robust coverage of rare events, long-horizon adaptation, or population-level variation. Future work should expand the scenario set (including harder edge cases), broaden the range of environments and interlocutors, and complement expert judgments with studies involving novice users.

\paragraph{Engineering completeness and system bugs.}
Our prototype still contains occasional bugs and rough edges typical of end-to-end real-time embodied pipelines (e.g., intermittent perception failures, tooling misfires, or timing artifacts). Our goal is not to claim a perfectly engineered product-ready system, but to demonstrate how real-time multimodal inference, tool calling, and robot control \emph{interplay} to produce or break grounded interaction. Nevertheless, these imperfections might have confounded our evaluation as failures can cascade across components. 


\section{Conclusion} 
\label{sec:conclusion}
This paper proposed a simple system recipe for \emph{situated embodied conversation}: a real-time multimodal language model that interleaves dialogue generation with tool-mediated attention and active perception. We instantiated this recipe on a social robot and studied it across six home-style scenarios. Using controlled ablations and backend comparisons, we evaluated both objective next-step tool decisions and subjective perceptions of interaction quality. Overall, the results suggest that the proposed system is a promising direction for situated embodied conversation in HRI. We hope this system recipe and evaluation protocol provide a concrete baseline for building and comparing future embodied conversation systems across environments and robot platforms.


\bibliographystyle{plainnat}
\bibliography{references}

\clearpage
\appendix 
\subsection{\textbf{Geometric binding layer: rigid transforms for mapping image targets to robot gaze commands}}
\label{app:geometric_binding}

To make egocentric video observations actionable, we explicitly \emph{bind} each camera frame to the robot's instantaneous camera pose and field-of-view (FOV). At time $t$, the camera pose is represented by a rigid transform from the camera optical frame $\mathcal{F}_C$ to the robot frame $\mathcal{F}_R$,
\begin{equation}
{}^{R}\mathbf{T}_{C}(t)=
\begin{bmatrix}
{}^{R}\mathbf{R}_{C}(t) & {}^{R}\mathbf{p}_{C}(t)\\
\mathbf{0}^\top & 1
\end{bmatrix}
\in SE(3),
\label{eq:TRC}
\end{equation}
obtained from robot kinematics and/or IMU state estimation. This transform provides a geometric bridge between image-space measurements and robot-frame gaze commands: for any 3D point ${}^{C}\mathbf{x}$ expressed in camera coordinates, the corresponding robot-frame point is
\begin{equation}
{}^{R}\mathbf{x}(t) = {}^{R}\mathbf{R}_{C}(t)\,{}^{C}\mathbf{x} + {}^{R}\mathbf{p}_{C}(t).
\label{eq:point_transform}
\end{equation}

\vspace{0.25em}
\noindent\textbf{(a) From pixels/keypoints to camera rays.}
Given camera intrinsics $\mathbf{K}$ (or equivalently the horizontal/vertical FOV and image size), any pixel measurement $(u,v)$---e.g., a pose keypoint, a bounding-box center, or a segmentation mask centroid---defines a viewing ray in the camera frame. Using normalized image coordinates $(u_n,v_n)\in[0,1]^2$ (with $u_n$ increasing to the right and $v_n$ increasing downward), a pinhole model yields the (unnormalized) ray direction
\begin{equation}
\tilde{\mathbf{r}}_{C}=
\begin{bmatrix}
(2u_n-1)\tan(\mathrm{hfov}/2)\\
(2v_n-1)\tan(\mathrm{vfov}/2)\\
1
\end{bmatrix},
\qquad
\mathbf{r}_{C}=\frac{\tilde{\mathbf{r}}_{C}}{\|\tilde{\mathbf{r}}_{C}\|}.
\label{eq:pixel_to_ray}
\end{equation}
This formulation makes FOV constraints explicit and supports direct reasoning about whether a hypothesized target lies within the camera frustum.

\vspace{0.25em}
\noindent\textbf{(b) Angular error from image-center offset.}
For closed-loop attentional control, we interpret the deviation of the target ray from the optical axis as an angular error. Let $\mathbf{r}_{C}=(r_x,r_y,r_z)^\top$ be the unit ray in $\mathcal{F}_C$, where the optical axis is $+\hat{\mathbf{z}}_C$. The yaw and pitch offsets required to rotate the optical axis onto the target ray are
\begin{equation}
\Delta\mathrm{yaw}=\arctan2(r_x,r_z), 
\qquad
\Delta\mathrm{pitch}=-\arctan2(r_y,r_z),
\label{eq:angular_error}
\end{equation}
where the negative sign on pitch follows the convention that image $v$ increases downward (targets below center correspond to ``look down''). Intuitively, $\Delta(\mathrm{yaw},\mathrm{pitch})$ provides a smooth, geometry-grounded error signal for recentering the target in view.

\vspace{0.25em}
\noindent\textbf{(c) Computing a robot-frame 3D look-at target.}
We use the rigid transform to express the desired gaze direction (and a corresponding 3D look-at point) in the robot frame so it can be sent to the robot's gaze controller. First, we rotate the camera ray into the robot frame:
\begin{equation}
\mathbf{r}_{R}(t) = {}^{R}\mathbf{R}_{C}(t)\,\mathbf{r}_{C}.
\label{eq:ray_rotate}
\end{equation}
We then parameterize a look-at point at a nominal distance $d$ along this direction from the camera origin:
\begin{equation}
{}^{R}\mathbf{g}(t) = {}^{R}\mathbf{p}_{C}(t) + d\,\mathbf{r}_{R}(t),
\label{eq:lookat_point}
\end{equation}
and publish $(x,y,z)={}^R\mathbf{g}(t)$ as the 3D gaze target. This construction remains well-defined even when metric depth is unavailable (we fix $d$, e.g., 1--2\,m); when depth estimates are available, $d$ can be replaced by an estimated range to produce a metric target.

\vspace{0.25em}
\noindent\textbf{(d) Shared closed-loop recentering for people and objects.}
Equations~\eqref{eq:angular_error}--\eqref{eq:lookat_point} define a generic image-servoing step that recenters an observed target in the egocentric view. In \texttt{look\_at\_person}, the target pixel $(u,v)$ is produced by a person-specific perception module (e.g., a face/nose keypoint from a pose tracker). In \texttt{look\_at\_object}, the target pixel is produced by an object-specific module (e.g., open-vocabulary detection/segmentation with temporal tracking), typically using a mask centroid or bounding-box center. In both cases, the controller converts the target's displacement from image center into an angular error $\Delta(\mathrm{yaw},\mathrm{pitch})$ and uses the bound pose ${}^R\mathbf{T}_C(t)$ to generate a robot-frame look-at goal $(x,y,z)$ that aligns the camera optical axis with the target. This shared geometric binding layer lets the same attentional control logic operate across people and objects, while keeping perception modular and interchangeable.

\end{document}